\documentclass[times,final,10pt]{article}

\usepackage{arxiv}
\usepackage{framed,multirow}

\usepackage{amssymb}
\usepackage{latexsym}

\usepackage{url}
\usepackage{xcolor}
\usepackage{times}
\usepackage{epsfig}
\usepackage{graphicx}
\usepackage{amsmath}
\usepackage{amssymb}
\usepackage{color}
\usepackage{subfig}
\usepackage{tablefootnote}
\usepackage{url}
\usepackage{breqn}
\usepackage{algorithm}
\usepackage{algorithmic}

\usepackage{graphicx}
\usepackage{amsmath}
\usepackage{amssymb}
\usepackage{booktabs}

\usepackage{multirow}

\def\eg{\emph{e.g}. } 

\def\ie{\emph{i.e}. } 
 
\def\etc{\emph{etc}. } 
 
\def\etal{\emph{et al}. }

\begin{document}

\title{DeepPoint3D: Learning Discriminative Local Descriptors using Deep Metric Learning on 3D Point Clouds}

\author{Siddharth Srivastava, Brejesh Lall \\ \ \\ Department of Electrical Engineering \\Indian Institute of Technology Delhi \\  eez127506@ee.iitd.ac.in, brejesh@ee.iitd.ac.in} 
\date{}
\maketitle


\begin{abstract}
Learning local descriptors is an important problem in computer vision. While there are many techniques for learning local patch descriptors for 2D images, recently efforts have been made for learning local descriptors for 3D points. The recent progress towards solving this problem in 3D leverages the strong feature representation capability of image based convolutional neural networks by utilizing RGB-D or multi-view representations. However, in this paper, we propose to learn 3D local descriptors by directly processing unstructured 3D point clouds without needing any intermediate representation. The method constitutes a deep network for learning permutation invariant representation of 3D points. To learn the local descriptors, we use a multi-margin contrastive loss which discriminates between similar and dissimilar points on a surface while also leveraging the extent of dissimilarity among the negative samples at the time of training. With comprehensive evaluation against strong baselines, we show that the proposed method outperforms state-of-the-art methods for matching points in 3D point clouds. Further, we demonstrate the effectiveness of the proposed method on various applications achieving state-of-the-art results.
\end{abstract}

\section{Introduction}

The ability to represent and uniquely match surface, regions and shape is a fundamental problem in computer vision. With 3D data, such problems are accompanied by many challenges in terms of size, complexity, quality and availability of sufficient data. The standard techniques for solving such a problem in 3D primarily involve keypoint identification, region alignment, construction of descriptors and  matching of descriptors. Further, the extraction of a descriptor may use information from only the keypoint (point descriptor), its neighbourhood (local descriptor), or the entire 3D model (global descriptor). The primary aim is to be able to robustly encode the properties of the region being described. The state-of-the-art methods for computing global descriptors in 3D are based on deep networks \cite{maturana2015voxnet, qi2016pointnet}. On the other hand, the most popular techniques for computing local descriptors are based on handcrafted techniques \cite{salti2014shot, guo2013rotational, guo20143d, guo2016comprehensive}. Recently, efforts have been made to leverage deep networks to learn local descriptors from 3D data \cite{khoury2017learning, huang2018learning} which derives motivation from similar efforts for 2D images \cite{zagoruyko2017deep}. However, these techniques do not work directly on the point cloud data, and instead, are based on extracting intermediate representations such as projections using multiple camera positions or histograms from the input 3D point cloud, that are passed into a deep network similar to a 2D convolutional neural network, from which the descriptor is obtained. Since these intermediate representations can result in loss of information, the resulting descriptor is not able to leverage the complete information available in the raw point cloud. Therefore, in this paper, we aim towards addressing this gap and using a deep network learn a local descriptor which can directly leverage from the point cloud representation of the 3D data. 

The point cloud representation can usually be obtained directly as an output of the devices such as Kinect. The advantage of such a point cloud is that it represents the closest approximation of the captured surface as compared to a voxelized cloud or a RGB-D image, and is comparatively easier to process \cite{zeng20173dmatch}. However, since such point clouds are unstructured \ie do not have any defined ordering of points, it is difficult to process them with traditional deep learning based methods. Many efforts have been made to process 3D data and extract meaningful information using the deep learning pipeline \cite{ioannidou2017deep}. Broadly, the deep networks that can process 3D data can be categorized into three types based on the type of input representation (i) Voxel based networks such as \cite{maturana2015voxnet} which take as input a voxelized cloud and apply 3D operations (convolution \textit{etc.}) in a pipeline similar to standard 3D Convolutional Neural Network (CNN) (ii) Networks based on standard 2D CNNs. There are two types of approaches with such networks. First, which take as input RGB and Depth images, and pass them into different networks prior to fusing them \cite{eitel2015multimodal}. And second, which compute multiple intermediate representations and pass them into separate 2D CNNs prior to combining the feature vectors \cite{khoury2017learning, su2015multi} (iii) The last and the most recent approach directly processes the unstructured point clouds, and usually involves learning a permutation invariant representation of the point cloud prior to further processing \cite{qi2016pointnet, klokov2017escape, qi2017pointnet++}. 

Since, the point clouds considered in this paper are unstructured, the deep learning techniques based on traditional convolutional operation cannot be applied. Therefore, we use deep networks that directly process unstructured 3D point clouds as our base network. The input to the network is a point along with its neighbourhood. We then employ a discriminative loss function to distinguish between good and bad matches. A drawback of such loss functions is that they give each pair of negative samples equal weightage, which makes the network saturate early or not learn at all in some cases. Therefore, we introduce a multi-margin contrastive loss to leverage the extent of separation amongst the negative samples themselves, \ie, we make the loss function incur a larger loss for hard negative examples, while a relatively lower loss is incurred for other (soft) negative examples. We compare the proposed technique against competing methods and strong baselines. Moreover, we show the generalization ability of the learned descriptors on several applications such as object classification, retrieval, binarization of descriptors, object discovery and drought stress classification of plants.

In view of the above, the contributions of this paper are 
\begin{itemize}
	\item We propose a multi margin contrastive loss function leading to more discriminative learning as compared to other loss functions. 
	\item We directly process 3D input point clouds with deep networks to learn local geometric descriptors.
	\item We construct a large scale 3D dataset for point correspondences. 
	\item We provide exhaustive experimental analysis and show that the proposed method outperforms the other competing deep learning based methods as well as traditional 3D descriptors.
\end{itemize}

The rest of the paper is organized as follows. Section \ref{sec:related} discusses various works related to classification and tag prediction. In Section \ref{sec:approach}, we describe the proposed methodology, while in Section \ref{sec:results}, the experimental setup and results are detailed. In Section \ref{sec:applications}, we show the application of the proposed descriptors on several practical problems. Finally, the conclusion is provided in Section \ref{sec:conclusion}.

\section{Related Works}\label{sec:related}
\subsection{Deep Learning based Global 3D Descriptors}
Shapenet from Wu \etal \cite{wu20153d} is a Convolutional Deep Belief Network for processing 3D data. The network has been used for shape completion and object recognition. Voxnet of Maturana \etal  \cite{maturana2015voxnet} takes as input a volumetric occupancy grid which can be constructed from LiDAR, RGB-D, Voxelized Clouds etc. The architecture for processing the volumetric occupancy grids comprises a 3D Convolutional Neural Network (3D-CNN) with a $128$ dimensional feature vector as output. Qi \etal
\cite{qi2016volumetric} evaluate various 3D-CNN architectures for volumetric data. They also introduce auxiliary learning tasks and long anisotropic kernels to improve object classification. Riegler \etal's \cite{riegler2016octnet} OctNet exploits the sparsity of 3D models by hierarchically partitioning them based on the density of the data. The convolutions are defined over these sub-partitions resulting in lower memory and computational requirements. This in turn enables the network to process inputs with higher resolutions as compared to other similar networks \cite{maturana2015voxnet, wu20153d}. Su \etal \cite{su2015multi} (MVCNN) use multiple views of 3D models as inputs to multiple CNNs while pooling the outputs in an end-to-end learning framework to obtain robust descriptors.

While the above approaches work with voxelized clouds or projections of 3D models, Qi \etal's PointNet \cite{qi2016pointnet} works directly with unstructured 3D point clouds by learning a symmetric function on the input data. The network can also learn point level features by aggregating descriptors within a segmentation network. The point features thus generated have been used for the task of classification as well \cite{srivastava2017drought}. Qi \etal also propose an extension of PointNet \ie PointNet++ \cite{qi2017pointnet++} that applies hierarchical learning to
PointNet for obtaining geometrically robust features. Klokov \etal \cite{klokov2017escape} propose
Kd-network where they construct a network of kd-trees for parameter learning and sharing. A major
advantage of these networks is that they avoid approximations introduced due to voxelization allowing them to capture fine details while also keeping the memory footprint low. Hence, in this work, we use these networks (specifically PointNet) to learn the local descriptors in 3D. 

\subsection{Deep Learning based Local 3D Descriptors}
As discussed above, there are several deep architectures for processing complete 3D models. However, recently researchers have made efforts to learn local descriptors as well from 3D data. This line of research seems similar to learning 2D patch descriptors \cite{zagoruyko2017deep, kumar2016learning}, but unlike images, the 3D models have varying input representations such as voxels, point clouds, meshes etc., which impact the choice of underlying architecture as well as the subsequent applications. With the same analogy, Huang \etal \cite{huang2018learning} describe a local 3D region by learning an embedding where geometrically and semantically similar points lie closer. They work on carefully selected projection of 3D views and use an image based convolutional neural network arranged as a siamese network with contrastive loss. Khoury \etal \cite{khoury2017learning} present an approach to learning features that represent the local geometry around a point in an unstructured point cloud. They construct a histogram by partitioning space around the point and providing as input, the resultant histogram, to a deep neural network mapping the histograms to low dimensional feature vectors. Zeng \etal \cite{zeng20173dmatch} learn local descriptors using a siamese network where the input are volumetric 3D patches from RGB-D reconstructions. Srivastava \etal \cite{srivastava2017large} learn features from supervoxels and use it for the task of object discovery. In contrast to these works, our method learns the local features directly from unstructured 3D point clouds.

\section{Methodology}\label{sec:approach}

\subsection{Overview}

\begin{figure*}[h]
	\centering
	\includegraphics[width=0.9\textwidth, height=70mm]{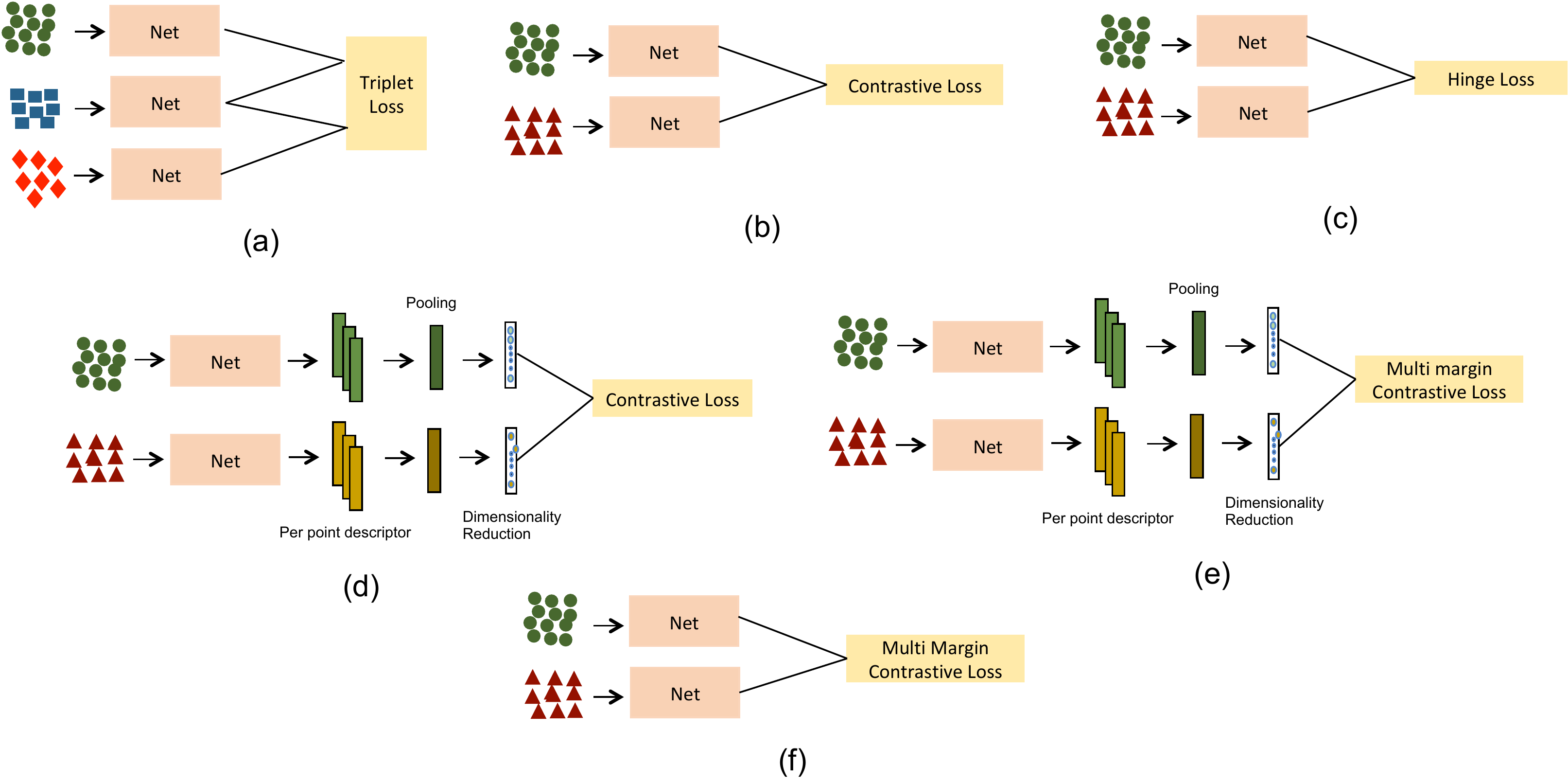}
	\caption{Proposed architectures for obtaining 3D local descriptors. (a) 3DPatch Triplet: A triplet Network with triplet loss consisting of Deep Point Cloud processing network (Net). The anchor samples are provided to the middle network, the positive samples are provided to the top network while the negative samples are provided to the bottom network. (b)-(c) 3DPatch Siamese: Siamese network consisting of deep network with Triplet loss, Contrastive Loss and Hinge Loss respectively. (d) Aggregated Siamese: Siamese network where per point feature from the deep network is pooled to a common descriptor followed by a fully connected layer for dimensionality reduction. (e)-(f) The proposed multi-margin contrastive loss function with both 3DPatch and Aggregated Siamese networks.}\label{fig:flow}
\end{figure*}

Figure \ref{fig:flow} shows the proposed architectures for extracting 3D point local descriptors. The input to the networks are regions around a 3D keypoint extracted using ISS keypoint detector \cite{zhong2009intrinsic}. The training set consists of pairs of positive and negative 3D patches. For multi-margin contrastive loss, the negative training samples are also labelled as hard negatives and simple negatives. This is motivated by the intuition that since the complexity of data varies it is feasible for hard examples to make the learning stronger by optimizing the constraint using a larger margin. The other examples can be used to tune the network as is done with traditional loss functions. Once the networks have been trained, it can be used to generate a $D$-dimensional feature vector as output by feeding the keypoint as input to the network. 
\subsection{Network Input}
The input to the network is a pair or a triplet of 3D patches. Each 3D local patch is a 3D keypoint extracted using ISS keypoint detector \cite{zhong2009intrinsic} and aligned using a Local Reference Frame \cite{salti2014shot} given as

\begin{equation} \label{eq:covmatrix}
\boldsymbol{M} = \frac{1}{\sum_{i:d_i \leq R}(R-d_i)} \sum_{i:d_i \leq R} (R-d_i)(\boldsymbol{p_i} - \boldsymbol{p})(\boldsymbol{p_i} - \boldsymbol{p})^T
\end{equation}

where $\boldsymbol{p}$ is keypoint in the 3D space, $R$ is the radius of the support region around the keypoint, $\boldsymbol{p_i}$ is a 3D point within a distance $R$ from the keypoint, and $d_i = ||\boldsymbol{p_i} - \boldsymbol{p}||_2$.

Once the surface is aligned, the standard method of defining the neighbourhood is to consider all points within a radius of $R$ from the keypoint. However, the resulting regions consists of variable number of points. While PointNet can handle variable number of points, we also consider the possibility of providing a fixed number of points as inputs which can represent the properties of the surrounding regions. In order to achieve this, we the method recently proposed in \cite{srivastava20163d} where the authors define the neighbourhood by selecting points separated by a certain angle in the 3D space(angular constraint) within a radius $R$. Therefore, the local region around the keypoint (3D patch) consists of $N$ neighbours of the keypoint satisfying the angular constraint. The choice of neighbourhood is important as it captures the geometrical property around the point while also impacts the computational complexity. In our experiments, we found the angular constraint based method to perform better than considering all the points in a neighbourhood. Therefore, for all the experiments in Section \ref{sec:results}, we report the results with neighbourhood constructed with angular constraint. 

\subsection{Loss Function}
Various losses have been evaluated. These losses are described next, followed by the proposed Multi-Margin Contrastive Loss. 
\\
\textbf{Hinge Loss}: Hinge loss is aimed towards maximum-margin classification. It is given as
\begin{align}
& \L (\Theta, x_i, x_j, y_{ij}) = \sum \left[ b - y_{ij} (m - \| f_\Theta (x_i) -  f_\Theta (x_j) \|^2)
\right]_+ \nonumber \\
& \textrm{where, } [a]_+ = \max(a,0) \; \forall a\in \mathbb{R}. 
\end{align}

The forward pass of the network is parameterized by $\Theta$ and is denoted as $f_\Theta(\cdot)$. $(x_i, x_j)$ is the pair of 3D patches provided as input to the network and $y_{ij} = +1$ or $-1$ indicates if they belong to the set positive pairs or negative pairs. $b \in \mathbb{R}$ and $m \in \mathbb{R}^+$ represent the hyper-parameters bias and margin respectively. 

\textbf{Contrastive Loss}:  The contrastive loss works by penalizing descriptors of positive pairs having large distance or descriptors of negative pairs having small distances. It is mathematically given as 
\begin{equation}
 \L (\Theta, x_i, x_j, y_{ij}) = y_{ij} * ||f_\Theta(x_i) - f_\Theta(x_j)||^2 +  (1-y_{ij})*max(0, m^2 - ||f_\Theta(x_i) - f_\Theta(x_j)||^2 )
\end{equation}

where $y_{ij} = +1$ for positive pairs and $0$ for negative pairs. $x_i, x_j$ is the pair of input 3D patches to the network.

\textbf{Triplet Loss}: A triplet loss computes the loss between an anchor, a positive and a negative patch with respect to the anchor patch. It aims to separate the positive and negative pairs by a specified distance margin. 
The triplet loss is given as 
\begin{equation}
\L (\Theta, x, x_{-}. x_{+}, y_{ij}) = max (0, 1-\frac{||f(x) - f(x_{-})||^2}{||f(x)-f(x_{+})||^2+m})
\end{equation}

where $x$, $x_{-}$, and $x_{+}$ are anchor, negative and positive samples respectively while $m$ is the margin. 

\textbf{Multi Margin Contrastive Loss}
While the above loss functions are popular, they have various drawbacks when the data is too complicated (no convergence) or too simple (saturation). Therefore, we introduce a multiple margins to the contrastive loss which can leverage the variety in the training data and therefore the network can be made to learn as per the complexity of the data instead of treating all examples as the same. The proposed loss function consists of two margins $m_1$ and $m_2$, where each margin corresponds to a particular set of training examples in the negative set. The negative training examples are categorized as soft negatives (simple training examples) or hard negatives (difficult training examples). The margin corresponding to hard negatives is kept higher than that for soft negatives since they have higher potential of providing discriminative information to assist the learning. This is mathematically given as 
\begin{multline}
\L (\Theta, x_i, x_j, y_{ij}, \gamma) = y_{ij} * ||f_\Theta(x_i) - f_\Theta(x_j)||^2 +  (1-y_{ij})*max(0, \gamma*({m^2_{1}} - ||f_\Theta(x_i) - f_\Theta(x_j)||^2),\\ (1-\gamma)*({m^2_{2}} - ||f_\Theta(x_i) - f_\Theta(x_j)||^2))
\end{multline}

where $y_{ij} = 1$ for positive pairs and $0$ for negative pairs. $x_i, x_j$ are input 3D patches. $\gamma$ is $1$ for hard negatives while $0$ for soft negatives. $m_1$ is the margin for hard negatives and $m_2$ is the margin of soft negatives. 

\subsection{Proposed Networks}
We have used PointNet and PointNet++ for directly processing the point clouds. To simplify the discussion, we would use \textit{Net} to refer to the deep network used in the proposed architectures as shown in Figure \ref{fig:flow}. Various types of networks proposed are:

\textbf{3DPatch triplet}: It consists of \textit{Nets} followed by a triplet loss function. The middle network is the anchor network while the first network is provided with the positive samples, and the third network is provided with the negative samples.

\textbf{3DPatch Siamese}: It consists of two \textit{Nets} with tied parameters followed by hingle, contrastive or multi-margin contrastive loss.

\textbf{Aggregated Siamese}: This is motivated by the fact that PointNet provides features using just a cloud point. Therefore, we aggregate the features for each of the points with pooling and train a fully connected layer on top of intermediate features. The output of the fully connected layer is forwarded to the loss function. The network has been evaluated with both Contrastive and Multi-Margin Contrastive Loss.
 
\subsection{Network Training}
The parameters in the proposed architecture are computed by minimizing the loss functions discussed previously. To avoid the large parameter values, we include a regularization term to the minimization as provided below

\begin{equation}
	\L_{R_t}(\mathbf{\chi}) = \L_t(\mathbf{\chi}) + \lambda||\mathbf{W}||^2
\end{equation}

where $\L_{R_t}$ is the regularized loss function, $\L_{t}$ is the loss function of type $t$, \ie, Hinge Loss, Contrastive Loss, Triplet Loss or Multi-Margin Contrastive Loss while $\mathbf{\chi}$ represents the corresponding parameters of the loss function $t$. $\lambda$ is the regularization parameter and $\mathbf{W}$ represents the weight vector.  

Additionally, the initialization for the \textit{Nets} has been done with the network trained for the task of part segmentation \cite{qi2016pointnet} on ShapeNet dataset.

\section{Experiments} \label{sec:results}

\subsection{Dataset}
We construct our dataset from the Shapenet Core dataset \cite{shapenet2015} with nearly $17,000$ shapes from $16$ categories. The dataset consists of labelled corresponding parts across various segmented 3D models. Therefore, we construct a set of corresponding pairs using the part-based registration from \cite{huang2018learning}. Further, we extract ISS keypoints \cite{zhong2009intrinsic} from the resultant parts and add to dataset, the points which have a corresponding keypoint in the corresponding pairs obtained previously. We construct the training set from $80$\% of the models, while $20$\% of the models are used for testing. For training the network, we need a set of positive and negative pairs. Therefore, to construct the set of positive pairs we add the matching keypoints from corresponding pairs across all the models. The negative pairs are constructed using two approaches, the first yields soft negatives, which are easier to distinguish, while the second results in hard negatives which are difficult to distinguish. We now describe both these approaches. The soft negatives are constructed by computing SHOT \cite{salti2014shot} descriptors for the points in the dataset and selecting points from distinct parts (same or different 3D models) which have a normalized Euclidean distance higher than a threshold. The intuition is that such points are keypoints (distinctive in their surroundings) and lie at a certain minimum distance in Euclidean space, they should be easy to distinguish in the embedding space as well. For hard negatives, we consider the neighbouring parts of same or similar 3D objects. For each such part, we compute the Nearest Neighbour Distance Ratio (NNDR) of the keypoints. The keypoints considered as a match are added to the set of hard negatives. Additionally, to introduce variation to the dataset, we randomly sample parts from different 3D models, and add the points which are a match as per NNDR criteria. However, as observed by \cite{srivastava2017large}, providing data at multiple resolutions augments the learning process from local 3D patches. Therefore, we sample point clouds at various mesh resolutions and add the resulting 3D patches to the dataset using the same methodology as described previously. 

\subsection{Evaluation Metrics}
The results are provide on four metrics, \ie Precision, Recall, Correspondence Accuracy and Cumulative Match Characteristics (CMC). These are explained below

\begin{itemize}
	\item Precision: It is given as the percentage of total matches which matched correctly to the ground-truth. Precision can be formulated as
	\begin{equation}
	Precision = \frac{\#Correct Matches}{\#Total Matches}
	\end{equation}
	\item Recall: It is given as the correct matches out of the total matches estimated as correct by the proposed technique. Recall is formulated as
	\begin{equation}
	Recall = \frac{\#Correct Matches}{\#Corresponding Matches}
	\end{equation}
	\item Cumulative Match Characteristics (CMC): For each pair of  shapes and input keypoint, we obtain a ranked list of matching points from one shape to the other shape as per the Euclidean distance amongst the descriptors of the points. Then we compute the fraction of ground-truth correspondences whose rank is less than $k$ retrieved matches from the target shape. We report the average over all the pair of shapes for all the keypoints.
	\item Correspondence Accuracy: For each pair of shapes, and input keypoint, the nearest point is found from one shape to the other shape. Then we calculate the Euclidean distance between the 3D position of the matched point and the ground-truth point. Finally, we compute the fraction of matched correspondences below a Euclidean error threshold $\tau$.
\end{itemize}

Several 3D models in the dataset have symmetric shapes, such as wingtips of airplanes. In such cases, for a feature point from one of the wingtips can be considered a match for points on wingtips on either side. Since, we have constructed the dataset from part labels of the shapes, such matches may present challenge in certain scenarios. Therefore, to study the impact of models with symmetric points, we report results by considering both symmetric and non-symmetric matches.

\subsection{Parameter Settings and Evaluation Setup}
For comparing against the techniques of LMVCNN \cite{huang2018learning}, CGF \cite{khoury2017learning}, 3DMatch \cite{zeng20173dmatch}, PCA \cite{kim2013learning}, Spin Image \cite{johnson1999using}, SHOT \cite{salti2014shot}, Shape Context \cite{kalogerakis2010learning} and SDF \cite{shapira2010contextual} we use the publicly available code provided by the authors and those available in Point Cloud Library. For experiments using Hinge Loss, Contrastive Loss and Triplet Loss, we set $m=1$, while for MMCL, we use $m_{1}=2$ and $m_{2}=1$. The dataset of soft negatives and hard negatives is merged to a single negative set for evaluation of architectures not using MMCL. Following the protocol of \cite{huang2018learning}, we also report result on training with single category and multiple category shapes. For single category training (Single Class), we train the network with the training correspondences obtained from a single class. For multiple category training, we train using correspondences from all $16$ classes of ShapeNetCore (Mixed 16). Additionally, for fair comparison with competing work, we also report results when the network is trained upon $13$ classes of ShapeNetCore by excluding airplanes, bikes, and chairs as reported by \cite{huang2018learning}. For reporting results with CMC and Correspondence Accuracy, we set the values of $k$ as $100$ and $\tau$ as $0.25$. 

\begin{table*}[]
	\centering
	\caption{Performance of the proposed architectures on the test dataset. The best performing method for each category of shape is shown in bold. }
	\label{tab:tab1}
	\footnotesize
	\resizebox{\textwidth}{!}{
		\begin{tabular}{|l|c|c|c|c|c|c|c|c|}
			\hline
			\multicolumn{1}{|c|}{\textbf{Model}}                                                                               & \textbf{\begin{tabular}[c]{@{}c@{}}CMC\\ (Symmetry)\end{tabular}} & \textbf{\begin{tabular}[c]{@{}c@{}}CMC\\ (No Symmetry)\end{tabular}} & \textbf{\begin{tabular}[c]{@{}c@{}}Corr. Acc.\\ (Symmetry)\end{tabular}} & \textbf{\begin{tabular}[c]{@{}c@{}}Corr. Acc.\\ (No Symmetry)\end{tabular}} & \textbf{\begin{tabular}[c]{@{}c@{}}Precision\\ (Symmetry)\end{tabular}} & \textbf{\begin{tabular}[c]{@{}c@{}}Precision\\ (No Symmetry)\end{tabular}} & \textbf{\begin{tabular}[c]{@{}c@{}}Recall\\ (Symmetry)\end{tabular}} & \textbf{\begin{tabular}[c]{@{}c@{}}Recall\\ (No Symmetry)\end{tabular}} \\ \hline
			\begin{tabular}[c]{@{}l@{}}3DPatch Siamese \\ (Hinge Loss) - Single Class\end{tabular}                             & 74.1                                                              & 70.1                                                                 & 48.5                                                                     & 44.4                                                                        & 0.75                                                                    & 0.71                                                                       & 0.71                                                                 & 0.61                                                                    \\ \hline
			\begin{tabular}[c]{@{}l@{}}3DPatch Siamese \\ (Contrastive Loss) - Single Class\end{tabular}                       & 77.5                                                              & 73.4                                                                 & 56.1                                                                     & 53.1                                                                        & 0.81                                                                    & 0.73                                                                       & 0.82                                                                 & 0.76                                                                    \\ \hline
			3DPatch Triplet - Single Class                                                                                     & 81.3                                                              & 75.5                                                                 & 61.1                                                                     & 58.8                                                                        & 0.84                                                                    & 0.80                                                                       & 0.83                                                                 & 0.75                                                                    \\ \hline
			\begin{tabular}[c]{@{}l@{}}Aggregated Siamese\\ (Contrastive Loss) - Single Class\end{tabular}                     & 73.5                                                              & 69.2                                                                 & 49.2                                                                     & 41.1                                                                        & 0.72                                                                    & 0.66                                                                       & 0.57                                                                 & 0.51                                                                    \\ \hline
			\begin{tabular}[c]{@{}l@{}}Aggregated Siamese\\ (Multi Margin Contrastive Loss)\\ Single Class\end{tabular}        & 76.4                                                              & 70.1                                                                 & 52.2                                                                     & 45.2                                                                        & 0.78                                                                    & 0.72                                                                       & 0.69                                                                 & 0.60                                                                    \\ \hline
			\begin{tabular}[c]{@{}l@{}}3D Patch Siamese\\ (Multi Margin Contrastive Loss)\\ Single Class\end{tabular} & \textbf{89.5}                                                     & \textbf{84.4}                                                        & \textbf{67.2}                                                            & \textbf{60.1}                                                               & \textbf{0.91}                                                           & \textbf{0.84}                                                              & \textbf{0.90}                                                        & \textbf{0.84}                                                           \\ \hline \hline
			\begin{tabular}[c]{@{}l@{}}3DPatch Siamese\\ (Hinge Loss) - Mixed 16\end{tabular}                                  & 69.1                                                              & 65.1                                                                 & 43.1                                                                     & 39.1                                                                        & 0.69                                                                    & 0.63                                                                       & 0.65                                                                 & 0.61                                                                    \\ \hline
			
			\begin{tabular}[c]{@{}l@{}}3DPatch Siamese \\ (Contrastive Loss) -  Mixed 16\end{tabular}                          & 72.2                                                              & 69.2                                                                 & 52.4                                                                     & 49.1                                                                        & 0.75                                                                    & 0.71                                                                       & 0.75                                                                 & 0.72                                                                    \\ \hline
			3DPatch Triplet - Mixed 16                                                                                         & 74.2                                                              & 72.2                                                                 & 55.3                                                                     & 52.2                                                                        & 0.76                                                                    & 0.72                                                                       & 0.75                                                                 & 0.72                                                                    \\ \hline
			\begin{tabular}[c]{@{}l@{}}Aggregated Siamese\\ (Contrastive Loss) - Mixed 16\end{tabular}                         & 68.3                                                              & 66.2                                                                 & 41.2                                                                     & 37.1                                                                        & 0.63                                                                    & 0.57                                                                       & 0.49                                                                 & 0.46                                                                    \\ \hline
			\begin{tabular}[c]{@{}l@{}}Aggregated Siamese\\ (Multi Margin Contrastive Loss)\\ Mixed 16\end{tabular}            & 74.4                                                              & 68.1                                                                 & 49.6                                                                     & 41.2                                                                        & 0.71                                                                    & 0.65                                                                       & 0.61                                                                 & 0.52                                                                    \\ \hline
			\begin{tabular}[c]{@{}l@{}}3D Patch Siamese\\ (Multi Margin Contrastive Loss)\\ \textit{Mixed 16}\end{tabular}    & \textbf{84.6}                                                     & \textbf{83.2}                                                        & \textbf{61.1}                                                            & \textbf{53.2}                                                               & \textbf{0.86}                                                           & \textbf{0.83}                                                              & \textbf{0.84}                                                        & \textbf{0.82}                                                           \\ \hline
		\end{tabular}
	}
	
\end{table*}

\subsection{Discussions}

\paragraph{Quantitative Results} In Table \ref{tab:tab1}, the results on the various proposed architectures are shown. It can be seen that 3D Patch Siamese with Multi Margin Contrastive Loss (MMCL) outperforms the next best candidate, Triplet Loss by nearly $8$\%, $6$\%, $10$\% and $11$\% respectively on CMC, Correspondence Accuracy, Precision and Recall (symmetric) for \textit{Single} Class. Similarly, 3D Patch Siamese (MMCL) outperforms the other proposed architectures when trained on \textit{Mixed 16} classes as well. It can be seen that the network with triplet loss performs better than other siamese based approaches (except MMCL). This means that having more discriminative information aids learning of better descriptors. As 3D Patch Siamese with MMCL performs best amongst the proposed architectures, in the subsequent experiments we provide results and comparisons against the 3D Patch Siamese (MMCL) and refer to it as 'Ours' throughout the text. Table \ref{tab:cmccorr} compares the matching performance of the best performing model amongst the proposed architectures, \ie 3D Patch Siamese (MMCL) against other deep and hand-crafted local descriptors. It can be observed that the proposed method outperforms all the other methods, with nearly $3\sim5\%$ improvement on an average on all metrics over LMVCNN, the second best performing method amongst the evaluated techniques. This shows that the proposed network can effectively encode the structure from the input 3D patch, while being more robust to symmetric examples.

\paragraph{Ablation Studies} To further demonstrate the effectiveness of the proposed method against noisy and low resolution 3D models, we perform experiments by varying the mesh resolution (mr) of the input 3D models. The results are shown in Tables \ref{tab:cmccorrmr0.25}, \ref{tab:cmccorr0.5}, and \ref{tab:cmccorr0.75}. It can be observed that as the resolution increases from $0.25$ to original (Table \ref{tab:cmccorr}), the matching performance increases. We also noticed that on low resolutions, the orientation assignment may become noisy, however, in our experiments we did not find any significant difference on low resolutions by avoiding normal computation while at high resolutions, the orientation assignment resulted in a significant boost in the matching performance. It can also be noted that on low resolutions ($0.25$, $0.5$) there is a significant drop in the performance of LMVCNN as compared to the same on higher resolutions. This is because LMVCNN uses projection of the point clouds to train the network and at low resolution, the generated projections become inaccurate inhibiting its performance. Similar trend is also observed for CGF. In other words, at lower resolution, the approaches based on intermediate representations \ie projections or histograms, suffer from significant information loss while the proposed technique can exploit the available information (geometry \etc) directly from the point cloud itself instead of any intermediate representation.

\begin{figure}[t]
	\centering
	\includegraphics[width=0.3\textwidth]{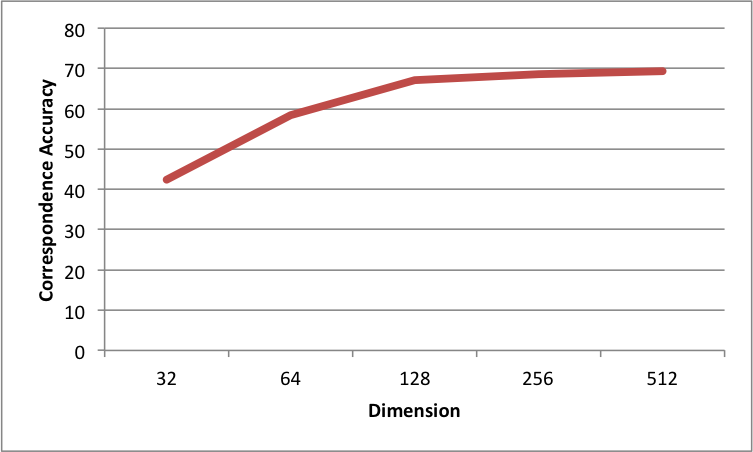} 
	\caption{
		Correspondence Accuracy on varying dimension of the descriptor with multi-margin contrastive loss.
	}
	\label{fig:dim}
\end{figure}

Figure \ref{fig:dim} shows the impact of varying the size of the descriptor obtained using the proposed technique. It can be seen that with low descriptor size (such as $32$, $64$) the correspondence accuracy is very low. This means that the learned embedding is not able to generalize well for these sizes, while CGF provides better results with $32$ bit vector. A possible reason is that in CGF, the learning is on the pre-processed patches where the network input is a histogram, while in the current setting the network input is the 3D patch itself. In our experiments, we found that making the network deeper (adding more layers), resulted in comparative performance with CGF having a descriptor size of $32$ (since the current model was able to generalize well on the evaluated settings, we report results with the default architecture.)

\paragraph{Qualitative Results}

\begin{figure}
	\centering
	\includegraphics[scale=0.35, height=62mm]{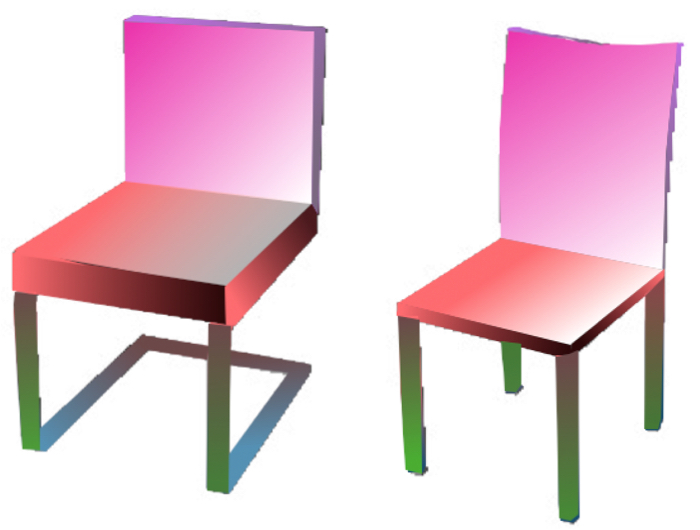}
	\caption{Visualization of local descriptors on two models of chair from ModelNet dataset}\label{fig:kpcomp}
\end{figure}

Figure \ref{fig:kpcomp} shows the variation of the descriptors learned over models of chairs. The figure was plotted by following the strategy of \cite{khoury2017learning} \ie we extract descriptor of each point of the model and project it to a $3$-dimensional RGB color space using PCA. Similar color on the surface indicates matching points in the two models. It can be observed that the correspondence is not only consistent but can also discriminate among various parts of the chairs such as the seat, back support and legs as indicated by similarity of colors on these parts. This is interesting as the method has not been trained explicitly to distinguish among parts. This also indicates that the network is able to learn the geometric structure of the surface. Moreover, the ground base of the first chair (left) has a bluish color, while such a color is not present in the second chair (right) indicating the absence of similar part (second chair does not have a base). This further demonstrates the disciminative ability of the learned features using the proposed model.

\begin{table*}[]
	\centering
	\caption{Comparison with state of the art at default mesh resolution. Here 'Ours' refers to 3D Patch Siamese with Multi Margin Contrastive Loss. The results in bold indicate the best performing method on the respective training sets. The results in \textit{italics} show the best performing hand-crafted descriptor. The overall best performing method for each category of shape is shown in bold.}
	\label{tab:cmccorr}
	\footnotesize
	\scalebox{0.8}{
	\begin{tabular}{|l|c|c|c|c|}
		\hline
		\multicolumn{1}{|c|}{\textbf{Method/Metric}}                     & \textbf{\begin{tabular}[c]{@{}c@{}}CMC\\  (Symmetry)\end{tabular}} & \textbf{\begin{tabular}[c]{@{}c@{}}CMC\\ (No Symmetry)\end{tabular}} & \textbf{\begin{tabular}[c]{@{}c@{}}Corr. Accuracy \\ (Symmetry)\end{tabular}} & \textbf{\begin{tabular}[c]{@{}c@{}}Corr. Accuracy\\ (No Symmetry)\end{tabular}} \\ \hline
		PCA \cite{kim2013learning}                                                             & 42.1                                                               & 38.5                                                                 & 38.6                                                                          & 31.8                                                                            \\ \hline
		SI \cite{johnson1999using}                                                               & 50.6                                                               & 48.2                                                                 & 47.9                                                                          & 41.1                                                                            \\ \hline
		SHOT \cite{salti2014shot}                                                           & 47.2                                                               & 44.6                                                                 & 43.1                                                                          & 40.6                                                                            \\ \hline
		SC \cite{kalogerakis2010learning}                                                              & \textit{76.3}                                                               & \textit{73.6}                                                                 & \textit{54.5}                                                                          & \textit{48.1}                                                                            \\ \hline
		SDF \cite{shapira2010contextual}                                                             & 33.8                                                               & 35.1                                                                 & 34.8                                                                          & 27.8   
		\\ \hline   
		FPFH \cite{rusu2009fast}                                                             & 37.5                                                               & 38.4                                                                 & 36.1                                                                          & 29.4        
		    \\ \hline   
		    RoPS\cite{guo2013rotational}                                                               & 48.1                                                               & 46.2                                                                 & 44.5                                                                          & 41.1                                                                    \\ \hline \hline
		\begin{tabular}[c]{@{}l@{}}LMVCNN \cite{huang2018learning}\\ (Single Class)\end{tabular}  & 84.3                                                               & 81.6                                                                 & 63.4                                                                          & 55.2                                                                            \\ \hline
		\begin{tabular}[c]{@{}l@{}}CGF \cite{khoury2017learning} \\ (Single Class)\end{tabular}    & 82.1                                                               & 79.1                                                                 & 60.3                                                                          & 53.3                                                                            \\ \hline
		\begin{tabular}[c]{@{}l@{}}Ours \\ (Single class)\end{tabular}   & \textbf{89.5}                                                               & \textbf{84.4}                                                                 & \textbf{67.2}                                                                          & \textbf{60.1}                                                                            \\                                                                  \hline \hline
		\begin{tabular}[c]{@{}l@{}}LMVCNN \cite{huang2018learning}\\ (Mixed 13)\end{tabular}      & 81.6                                                               & 74.3                                                                 & 53.2                                                                          & 43.1                                                                            \\ \hline
		\begin{tabular}[c]{@{}l@{}}CGF \cite{khoury2017learning} \\ (Mixed 13)\end{tabular}        & 77.3                                                               & 74.3                                                                 & 50.7                                                                          & 40.1                                                                            \\ \hline
		\begin{tabular}[c]{@{}l@{}}Ours\\ (Mixed 13)\end{tabular}        & \textbf{84.1}                                                              & \textbf{78.4}                                                                 & \textbf{55.1}                                                                          & \textbf{47.3}                                                                            \\ \hline \hline
		\begin{tabular}[c]{@{}l@{}}LMVCNN \cite{huang2018learning}\\ (Mixed 16)\end{tabular}      & 83.1                                                               & 80.5                                                                 & 57.6                                                                          & 49.6                                                                            \\ \hline
		\begin{tabular}[c]{@{}l@{}}CGF \cite{khoury2017learning}\\ (Mixed 16)\end{tabular}         & 79.1                                                               & 77.2                                                                 & 54.4                                                                          & 46.5                                                                            \\ \hline
		\begin{tabular}[c]{@{}l@{}}Ours\\ (Mixed 16)\end{tabular}        & \textbf{84.6}                                                               & \textbf{83.2}                                                                 & \textbf{61.1}                                                                         & \textbf{53.2}
		\\ \hline
	\end{tabular}
}
\end{table*}

\begin{table*}[]
	\centering
	\caption{CMC and Correspondence Accuracy on test dataset at mr = 0.25 for deep network based methods. Here 'Ours' refers to 3D Patch Siamese with Multi Margin Contrastive Loss}
	\label{tab:cmccorrmr0.25}
	\footnotesize
	\scalebox{0.8}{
	\begin{tabular}{|l|c|c|c|c|}
		\hline
		\multicolumn{1}{|c|}{\textbf{Method/Metric}}                    & \textbf{\begin{tabular}[c]{@{}c@{}}CMC\\  (Symmetry)\end{tabular}} & \textbf{\begin{tabular}[c]{@{}c@{}}CMC\\ (No Symmetry)\end{tabular}} & \textbf{\begin{tabular}[c]{@{}c@{}}Corr. Accuracy \\ (Symmetry)\end{tabular}} & \textbf{\begin{tabular}[c]{@{}c@{}}Corr. Accuracy\\ (No Symmetry)\end{tabular}} \\ \hline
		\begin{tabular}[c]{@{}l@{}}LMVCNN\\ (Single Class)\end{tabular} & 51.1                                                               & 49.2                                                                 & 36.2                                                                          & 33.9                                                                            \\ \hline
		\begin{tabular}[c]{@{}l@{}}CGF \\ (Single Class)\end{tabular}   & 53.5                                                               & 52.4                                                                 & 38.4                                                                          & 35.8                                                                            \\ \hline 
			\begin{tabular}[c]{@{}l@{}}Ours \\ (Single Class)\end{tabular}  & \textbf{67.1}                                                               & \textbf{65.1}                                                                 & \textbf{43.2}                                                                          & \textbf{41.1}                                                                            \\ \hline \hline
		\begin{tabular}[c]{@{}l@{}}LMVCNN\\ (Mixed 13)\end{tabular}     & 45.1                                                               & 41.5                                                                 & 33.6                                                                          & 30.8                                                                            \\ \hline
		\begin{tabular}[c]{@{}l@{}}CGF \\ (Mixed 13)\end{tabular}       & 48.1                                                               & 43.4                                                                 & 36.2                                                                          & 32.7                                                                            \\ \hline
		\begin{tabular}[c]{@{}l@{}}Ours\\ (Mixed 13)\end{tabular}       & \textbf{54.3 }                                                              & \textbf{52.3}                                                                 & \textbf{41.7}                                                                          & \textbf{39.3}                                                                            \\ \hline \hline
	
			\begin{tabular}[c]{@{}l@{}}LMVCNN\\ (Mixed 16)\end{tabular}     & 46.2                                                               & 42.6                                                                 & 34.4                                                                          & 31.3                                                                            \\ \hline
		
		\begin{tabular}[c]{@{}l@{}}CGF\\ (Mixed 16)\end{tabular}        & 48.9                                                               & 43.1                                                                 & 36.3                                                                          & 32.4                                                                            \\ \hline
		\begin{tabular}[c]{@{}l@{}}Ours\\ (Mixed 16)\end{tabular}       & \textbf{55.3}                                                               & \textbf{53.2}                                                                 & \textbf{42.2}                                                                         & \textbf{39.4}                                                                            \\ \hline
		
	\end{tabular}}
\end{table*}

\begin{table*}[]
	\centering
	\caption{CMC and Correspondence Accuracy on test dataset at mr=0.5 for deep network based methods. Here 'Ours' refers to 3D Patch Siamese with Multi Margin Contrastive Loss}
	\footnotesize
	\label{tab:cmccorr0.5}
	\scalebox{0.8}{
	\begin{tabular}{|l|c|c|c|c|}
		\hline
		\multicolumn{1}{|c|}{\textbf{Method/Metric}}                    & \textbf{\begin{tabular}[c]{@{}c@{}}CMC\\  (Symmetry)\end{tabular}} & \textbf{\begin{tabular}[c]{@{}c@{}}CMC\\ (No Symmetry)\end{tabular}} & \textbf{\begin{tabular}[c]{@{}c@{}}Corr. Accuracy \\ (Symmetry)\end{tabular}} & \textbf{\begin{tabular}[c]{@{}c@{}}Corr. Accuracy\\ (No Symmetry)\end{tabular}} \\ \hline
		\begin{tabular}[c]{@{}l@{}}LMVCNN\\ (Single Class)\end{tabular} & 55.3                                                               & 53.2                                                                 & 41.3                                                                          & 38.5                                                                            \\ \hline
		\begin{tabular}[c]{@{}l@{}}CGF \\ (Single Class)\end{tabular}   & 57.5                                                               & 55.3                                                                 & 43.5                                                                          & 41.3                                                                            \\ \hline
		\begin{tabular}[c]{@{}l@{}}Ours \\ (Single Class)\end{tabular}  & \textbf{71.1}                                                               & \textbf{67.8}                                                                 & \textbf{48.4}                                                                          & \textbf{43.4}                                                                            \\ \hline \hline
		\begin{tabular}[c]{@{}l@{}}LMVCNN\\ (Mixed 13)\end{tabular}     & 51.3                                                               & 47.2                                                                 & 38.5                                                                          & 33.5                                                                            \\ \hline
			\begin{tabular}[c]{@{}l@{}}CGF \\ (Mixed 13)\end{tabular}       & 54.3                                                               & 57.6                                                                 & 43.2                                                                          & 37.4                                                                            \\ \hline
			\begin{tabular}[c]{@{}l@{}}Ours\\ (Mixed 13)\end{tabular}       & \textbf{55.4}                                                               & \textbf{53.4}                                                                 & \textbf{43.5}                                                                          & \textbf{43.4}                                                                            \\ \hline \hline
				\begin{tabular}[c]{@{}l@{}}LMVCNN\\ (Mixed 16)\end{tabular}     & 51.6                                                               & 48.5                                                                 & 38.7                                                                          & 35.5                                                                            \\ \hline
		\begin{tabular}[c]{@{}l@{}}CGF\\ (Mixed 16)\end{tabular}        & 54.4                                                               & 58.5                                                                 & 43.4                                                                          & 38.5                                                                            \\ \hline
	
		\begin{tabular}[c]{@{}l@{}}Ours\\ (Mixed 16)\end{tabular}       & \textbf{58.1}                                                              & \textbf{57.3}                                                                 & \textbf{45.3}                                                                          & \textbf{43.2}                                                                            \\ \hline

	\end{tabular}}
\end{table*}

\begin{table*}[]
	\centering
	\caption{CMC and Correspondence Accuracy on test Dataset with mr = 0.75 for deep network based methods. Here 'Ours' refers to 3D Patch Siamese with Multi Margin Contrastive Loss}
	\label{tab:cmccorr0.75}
	\footnotesize
	\scalebox{0.8}{
	\begin{tabular}{|l|c|c|c|c|}
		\hline
		\multicolumn{1}{|c|}{\textbf{Method/Metric}}                    & \textbf{\begin{tabular}[c]{@{}c@{}}CMC\\  (Symmetry)\end{tabular}} & \textbf{\begin{tabular}[c]{@{}c@{}}CMC\\ (No Symmetry)\end{tabular}} & \textbf{\begin{tabular}[c]{@{}c@{}}Corr. Accuracy \\ (Symmetry)\end{tabular}} & \textbf{\begin{tabular}[c]{@{}c@{}}Corr. Accuracy\\ (No Symmetry)\end{tabular}} \\ \hline
		\begin{tabular}[c]{@{}l@{}}LMVCNN\\ (Single Class)\end{tabular} & 76.3                                                               & 74.5                                                                 & 55.6                                                                          & 47.3                                                                            \\ \hline
		\begin{tabular}[c]{@{}l@{}}CGF \\ (Single Class)\end{tabular}   & 73.4                                                               & 70.1                                                                 & 53.4                                                                          & 44.4                                                                            \\ \hline
		\begin{tabular}[c]{@{}l@{}}Ours \\ (Single Class)\end{tabular}  & \textbf{80.3}                                                               & \textbf{75.4}                                                                 & \textbf{61.1}                                                                          & \textbf{51.1}                                                                            \\ \hline \hline
	
		\begin{tabular}[c]{@{}l@{}}LMVCNN\\ (Mixed 13)\end{tabular}     & 75.3                                                               & 68.3                                                                 & 48.6                                                                          & 39.9                                                                            \\ \hline
		\begin{tabular}[c]{@{}l@{}}CGF \\ (Mixed 13)\end{tabular}       & 71.1                                                               & 69.3                                                                 & 43.3                                                                          & 39.4                                                                            \\ \hline
		\begin{tabular}[c]{@{}l@{}}Ours\\ (Mixed 13)\end{tabular}       & \textbf{78.9}                                                               & \textbf{71.3}                                                                 & \textbf{52.2}                                                                          & \textbf{43.2}                                                                            \\ \hline \hline
			\begin{tabular}[c]{@{}l@{}}LMVCNN\\ (Mixed 16)\end{tabular}     & 76.3                                                               & 67.5                                                                 & 46.3                                                                          & 40.1                                                                            \\ \hline
			\begin{tabular}[c]{@{}l@{}}CGF\\ (Mixed 16)\end{tabular}        & 71.1                                                               & 68.4                                                                 & 47.5                                                                          & 43.5                                                                            \\ \hline
		\begin{tabular}[c]{@{}l@{}}Ours\\ (Mixed 16)\end{tabular}       & \textbf{73.4}                                                               & \textbf{74.5}                                                                 & \textbf{53.5}                                                                          & \textbf{49.3}                                                                            \\ \hline
	\end{tabular}}
\end{table*}

\section{Applications}\label{sec:applications}
In this section, we show the performance of the proposed descriptors on various basic as well as advanced tasks. The basic tasks involve classification, retrieval \etc The base problems have been chosen is to show the improvements obtained using the proposed technique when it is used as a drop-in replacement for handcrafted features in traditional pipelines. The advanced tasks are more specific tasks directly solving a practical real world problem. 

\subsection{Classification}
Given an input 3D model, the objective is to identify the class to which the model belongs. To show the effectiveness of the proposed local descriptors against other local descriptors, we follow a standard pipeline of extracting keypoints from point clouds and describe them using various descriptors. Next, the descriptors are quantized using Fisher Vector \cite{sanchez2013image} and classified using a Support Vector Machine. The classification is performed on Semantic3D\cite{hackel2017isprs} and ModelNet10\cite{wu20153d} dataset. The Semantic3D provides point level labels, hence local descriptors are more suited for such classification. ModelNet10 does not provide point level labels, and hence in general, the global descriptors are preferred on ModelNet10 for evaluation. However, we provide results on ModelNet10 for two reasons. First, while it is amongst the most popular benchmarks, not many 3D local descriptors have been evaluated on it and hence, we provide a relative comparison of local descriptors on large 3D models and datasets. Second, since ModelNet10 does not provide point level labels, we use the network fine-tuned on Semantic3D for classification and still achieve better results than the compared local descriptors demonstrating the generalization ability of the proposed method.

For the results reported for classification, we use the pre-trained network from 3D Patch Siamese (MMCL) used in previous experiments, and fine tune it for classification task using the labelled training points provided by Semantic3D. The results are shown in Table \ref{tab:classification} and Table \ref{tab:classificationret} (a) for Semantic3D and ModelNet10 respectively. It can be seen that the proposed method outperforms the compared methods with LMVCNN lagging behind by $2$\% on Semantic3D and by $3.4$\% on ModelNet10. It can also be noticed that the deep descriptors consistently perform superior to the traditional hand-crafted descriptors \eg RoPS lags behind the proposed descriptor by $7$\% on Semantic3D. This indicates that the learning based methods are able to encode, the local surface information, in a more robust manner, where the proposed architecture outperforms the deep learning based methods by $\sim2.5\%$ on Semantic3D and $\sim3.8$\% on ModelNet10 on an average. The table also shows the results of a few state-of-the-art techniques on ModelNet10 dataset. These techniques provide global descriptors. It can be seen that the performance of state-of-the-art is greater than $95$\%, however, such techniques are usually based on alternate representations of point clouds such as projections \cite{yavartanoo2018spnet} or ensemble of networks \cite{sfikas2018ensemble} to achieve such results instead of directly processing point clouds. 

\begin{table}[]
	\centering
	\caption{Classification Accuracy on Semantic3D with local descriptors}
	\label{tab:classification}
	\footnotesize
	\scalebox{0.8}{
	\begin{tabular}{|c|c|}
		\hline
		\textbf{Technique} & \textbf{\begin{tabular}[c]{@{}c@{}}Accuracy\\ (\%)\end{tabular}} \\ \hline
		SHOT \cite{salti2014shot}              & 57.0                                                             \\ \hline
		RoPS  \cite{guo2013rotational}               & 61.1                                                             \\ \hline
		FPFH \cite{rusu2009fast}               & 42.3                                                             \\ \hline
		SI \cite{johnson1999using}               &   59.4                                                           \\ \hline
		PCA \cite{kim2013learning}               &    45.8                                                          \\ \hline
		SC \cite{kalogerakis2010learning}               &   61.5                                                           \\ \hline
		SDF \cite{shapira2010contextual}            &    33.2                                                         \\ \hline
		LMVCNN \cite{huang2018learning}            & 66.1                                                             \\ \hline
		CGF  \cite{khoury2017learning}              & 65.6                                                             \\ \hline
		Ours               & \textbf{68.1}                                                    \\ \hline
	\end{tabular}}

\end{table}

\begin{table}%
	\centering
	\caption{Classification and Retrieval performance on  ModelNet10 dataset with real valued descriptors (a) Classification on ModelNet10 (b) Retrieval on ModelNet10. The best performing local hand-crafted descriptor is shown in \textit{italics} while the overall best performing local descriptor is shown in bold.}
	\label{tab:classificationret}
	\subfloat[][]{
	\scalebox{0.65}{
		\begin{tabular}{|c|c|}
			\hline
			\textbf{Technique} & \textbf{\begin{tabular}[c]{@{}c@{}}Acc.(\%)\end{tabular}} \\ \hline
			\multicolumn{2}{|l|}{Local Descriptors (Hand-crafted)}                                                                 \\ \hline
			SHOT \cite{salti2014shot}              & 48.5                                                            \\ \hline
			RoPS  \cite{guo2013rotational}               &    51.6                                                          \\ \hline
			FPFH \cite{rusu2009fast}  
			&   40.8                                            \\ \hline
			SI \cite{johnson1999using}               &  55.4                                                            \\ \hline
			PCA \cite{kim2013learning}               &  45.6                                                            \\ \hline
			SC \cite{kalogerakis2010learning}               &  \textit{58.2}                                                            \\ \hline
			SDF \cite{shapira2010contextual}            & 28.3                                                            \\ \hline \hline
			\multicolumn{2}{|l|}{Local Descriptors (Deep Learning)} \\ \hline
			LMVCNN \cite{huang2018learning}            & 66.1                                                             \\ \hline
			CGF  \cite{khoury2017learning}              &   65.5                                                           \\ \hline
			Ours               & \textbf{69.5}                                                    \\ \hline \hline
			\multicolumn{2}{|l|}{Global Descriptors (Deep Learning)}            \\ \hline 
			RotationNet \cite{kanezaki2018rotationnet}                & 98.46  \\ \hline        
			SPNet \cite{yavartanoo2018spnet}                & 97.25                                                       \\ \hline
	\end{tabular}}
	\label{It_Ith1}
	}
	\qquad
	\subfloat[][]{
		\scalebox{0.65}{
			\begin{tabular}{|c|c|}
				\hline
				\textbf{Technique} & \textbf{mAP(\%)} \\ \hline
				\multicolumn{2}{|l|}{Local Descriptors (Hand-crafted)}                                                                 \\ \hline
				SHOT \cite{salti2014shot}                & 38.6             \\ \hline
				RoPS \cite{guo2013rotational}              & 41.3             \\ \hline
				FPFH \cite{rusu2009fast}               & 34.3             \\ \hline
				SI \cite{johnson1999using}               &   31.4                                                           \\ \hline
				PCA \cite{kim2013learning}               &    27.5                                                          \\ \hline
				SC \cite{kalogerakis2010learning}               & \textit{42.4}                                                             \\ \hline
				SDF \cite{shapira2010contextual}            &  25.6                                                           \\ \hline \hline
				\multicolumn{2}{|l|}{Local Descriptors (Deep Learning)} \\ \hline
				LMVCNN \cite{huang2018learning}            & 44.2             \\ \hline
				CGF \cite{khoury2017learning}               & 48.6             \\ \hline
				Ours               & \textbf{48.8}    \\ \hline \hline
				\multicolumn{2}{|l|}{Global Descriptors (Deep Learning)}            \\ \hline         
				SPNet \cite{yavartanoo2018spnet}                & 94.2                                                       \\ \hline
				PANORAMA-ENN \cite{sfikas2018ensemble}                & 93.28  \\ \hline
		\end{tabular}}
	}
\end{table}

\subsection{Retrieval}
Retrieval aims at fetching, usually from a large repository, models similar to the input model. For efficient retrieval, the features need to robustly encode local information to differentiate among similarly structured models. The retrieval pipeline involves extracting features from keypoints obtained from the 3D models, quantizing them using Fisher Vector and then matching them with the quantized descriptors from the models in the repository. For evaluation we use the ModelNet10 dataset \cite{wu20153d} and the results are shown in Table \ref{tab:classificationret}(b). It can be seen that the proposed method outperforms the compared methods by $\sim7$ on an average. Moreover, it significantly outperforms the handcrafted descriptors \eg RoPS lags behind by $7.5$\% while SHOT is outperformed by $10.2$\%. An interesting observation here is that CGF performs better than LMVCNN while only slightly lagging behind the proposed method. This is because CGF is specifically designed to robustly encode the geometric information, while LMVCNN and our method rely on general purpose encoding of descriptors. This indicates that the proposed descriptor can effectively encode the geometric information into a compact descriptor.

\subsection{Binarization}
Compactness of feature plays an important role in scaling up the number of models that can be evaluated for a specific objective. However, at the same time it is also important to maintain the robustness of the original descriptor. Therefore, we binarize the descriptors and evaluate their performance on  classification and retrieval tasks discussed above. The binarization for real-valued descriptors is achieved through Iterative Quantization \cite{gong2013iterative} where the descriptors are reduced to $128$ bits while CGF has been reduced to $32$ bit (as maximum number of real values in the descriptor is $32$). The results are shown in Table \ref{tab:bin}. It can be seen that the deep learning based descriptors consistently outperform handcrafted descritpors \ie on an average the handcrafted descriptors lag behind by $\sim20$\% from the proposed method. However, the 3D Binary Signatures \cite{srivastava20163d} performs closer to the deep learning based descriptors, where 3DBS-64 \ie 3DBS with $64$ nearest neighbours, lags behind by only $5.5$\%. This shows that even with significant reduction in the descriptor size, the performance on classification and retrieval tasks is nearly the same as those of their respective real valued versions (Table \ref{tab:classification}, Table \ref{tab:classificationret}).

\begin{table}[]
	\centering
	\caption{Classification and Retrieval Performance on binary descriptor. The real valued descriptors have been quantized to $128$ bits using Iterative Quantization ($32$ bits for CGF).}
	\label{tab:bin}
	\footnotesize
	\footnotesize
	\scalebox{0.85}{
	\begin{tabular}{|l|c|l|}
		\hline
		\textbf{Method} & \multicolumn{1}{l|}{\textbf{\begin{tabular}[c]{@{}l@{}}Classification\\ Accuracy (\%)\end{tabular}}} & \textbf{\begin{tabular}[c]{@{}l@{}}Retrieval\\ mAP (\%)\end{tabular}} \\ \hline
		\multicolumn{3}{|l|}{Local Descriptors (Hand-crafted)}                                                                 \\ \hline
		SHOT \cite{salti2014shot}            & 42.3                                                                                                 & 20.1                                                                  \\ \hline
		RoPS \cite{guo2013rotational}           & 43.0                                                                                                 & 24.8                                                                  \\ \hline
		FPFH \cite{rusu2009fast}           & 22.8                                                                                                 & 14.3                                                                  \\ \hline
		SI \cite{johnson1999using}               &   40.1 & 17.5                                                          \\ \hline
		PCA \cite{kim2013learning}               &    30.4         &  12.6                                               \\ \hline
		SC \cite{kalogerakis2010learning}               &   \textit{44.2}    &     \textit{24.1}                                                  \\ \hline
		SDF \cite{shapira2010contextual}            &    20.2   &   16.6                                                   \\ \hline
		3DBS-32 \cite{srivastava20163d}        & 54.0                                                                                                 & 34.1                                                                  \\ \hline
		3DBS-64 \cite{srivastava20163d}        & 57.6                                                                                                 & 38.1                                                                  \\ \hline
		B-SHOT \cite{prakhya2015b}  & 40.1                                                                                                 & 18.6                                                                                                                                \\ \hline
		\multicolumn{3}{|l|}{Local Descriptors (Deep Learning)}                                                                 \\ \hline
		CGF \cite{khoury2017learning}            & 59.8                                                                                                 & 41.1                                                                  \\ \hline
		LMVCNN \cite{huang2018learning}         & 62.1                                                                                                 & 40.4                                                                  \\ \hline
		Ours            & \textbf{63.1}                                                                                        & \textbf{41.4}                                                         \\ \hline
	\end{tabular}}
\end{table}

\subsection{Object Discovery in 3D}

\begin{figure}
	\centering
	
	\begin{tabular}{ccc}
		\includegraphics[width=24mm]{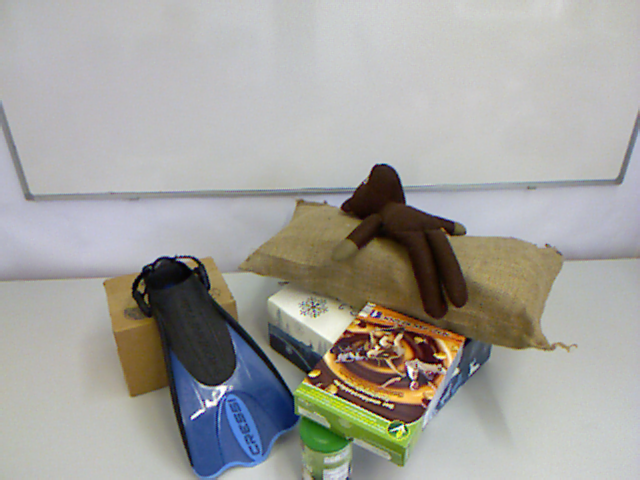} &   
		\includegraphics[width=24mm]{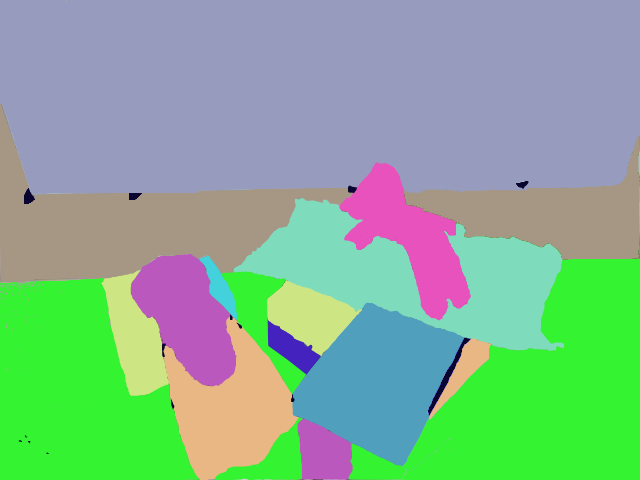} &
		\includegraphics[width=24mm]{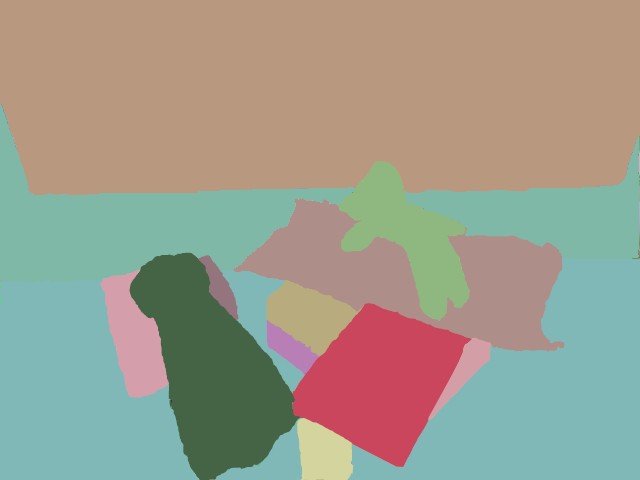} \\
		\includegraphics[width=24mm]{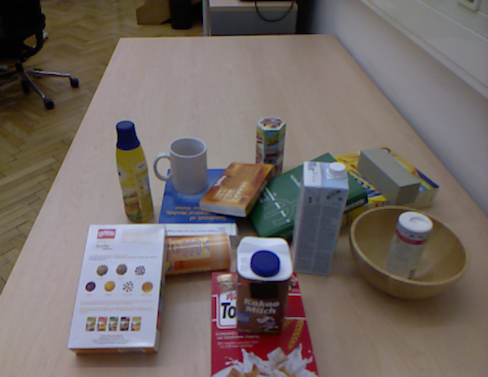} &  
		\includegraphics[width=24mm]{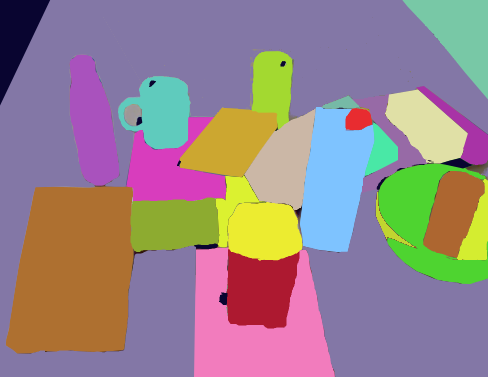} &
		\includegraphics[width=24mm]{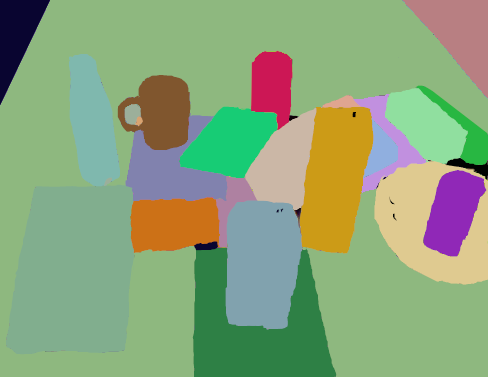}  \\
	\end{tabular}
	\caption{Qualitative results of discovering objects on various images. From the left, the first image is the original image, the second image is the result obtained using \cite{srivastava2017large} and the third image is the result obtained using our technique.}
	\label{fig:osdcumu} 
\end{figure}

\begin{figure*}
	\includegraphics[width=0.32\textwidth]{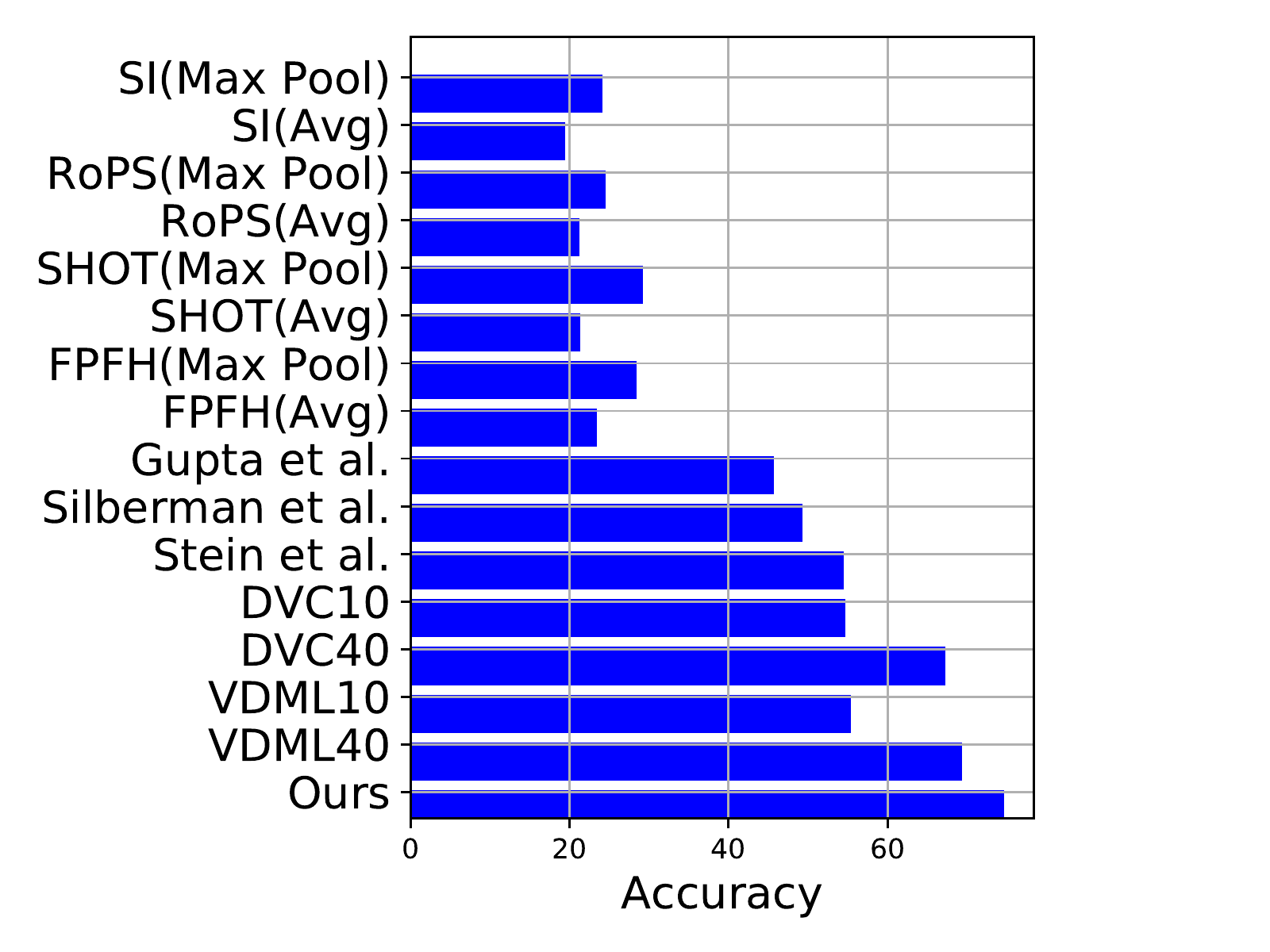} 
	\includegraphics[width=0.32\textwidth]{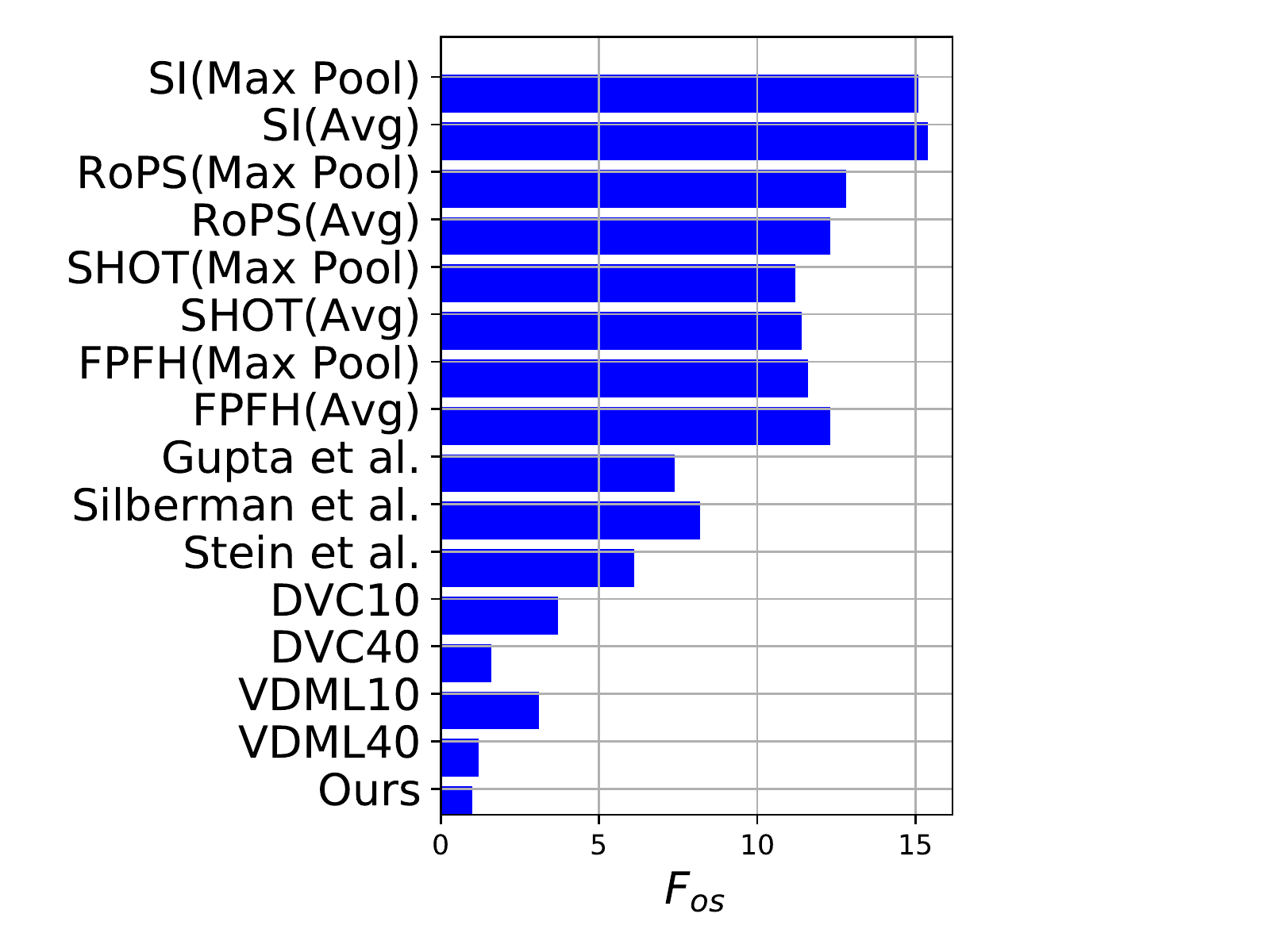}
	\includegraphics[width=0.32\textwidth]{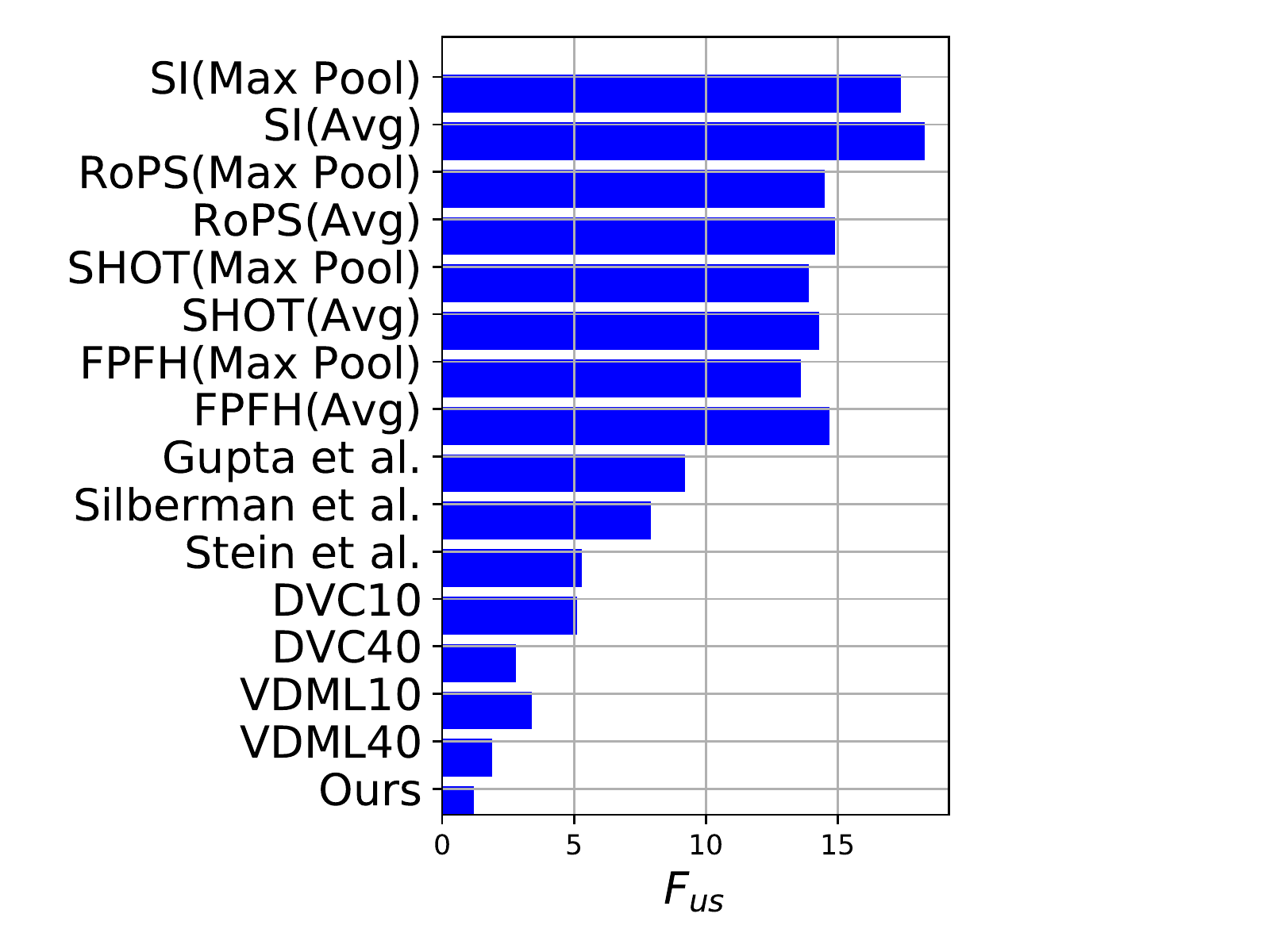}
	\caption{
		Accuracies, $F_{os}$ and $F_{us}$ scores on NYU Depth v2
		dataset. Lower is better for $F_{os}$ and $F_{us}$ scores.
	}
	\label{fig:objdisc}
\end{figure*}

Object Discovery in 3D refers to finding never-before-seen objects in 3D scenes. For discovering objects, we follow the pipeline proposed by \cite{srivastava2017large}. We use the pre-trained network proposed in this paper and fine-tune it on the \textit{Universal Training Set} proposed by the authors. However, our pipeline differs in the following way. Instead of a voxel based 3d convolutional neural network, we use the network proposed in this work. Therefore, instead of voxelizing the supervoxels, we use the corresponding points (unstructured) as input to the network. Additionally, instead of using a hinge loss, we use the multi-margin contrastive loss proposed in the paper. We show results on the NYU Depth v2 dataset \cite{silberman2012indoor} in Figure \ref{fig:objdisc}. The proposed method outperforms VDML40 by $2.3$\% while also having lower over-segmentation ($F_{os}$) and under-segmentation ($F_{us}$) scores as compared to variations of VDML, local descriptors (SI, RoPS, SHOT, FPFH) as well as those of Gupta \etal \cite{gupta2013perceptual}, Silberman \etal\cite{silberman2012indoor} and Stein \etal \cite{stein2014object}.  

Figure \ref{fig:osdcumu} shows qualitative results for object discovery using the proposed method on Object Segmentation Dataset \cite{richtsfeld2012segmentation} and Object Discovery Dataset\cite{mueller2016hierarchical} with a network pre-trained on NYU Depth v2. It can be observed that the proposed method provides better coverage of objects even with varying colors. For example, in first row, the vaccum cleaner is detected as two distinct objects (over-segmentation) by \cite{srivastava2017large}, while the proposed method classifies it as a single object. This can be attributed to the point based learning of neighbourhood information which allows it learn finer details within the objects, whereas, voxel based representation tends to loose such information. Similarly, in second row, the discovered objects have lesser over-segmentation for objects such as cup, bowl which have curved surfaces.

\subsection{Drought Stress Classification}

\begin{figure}[t]
	\centering
	\includegraphics[scale=0.45, height=55mm]{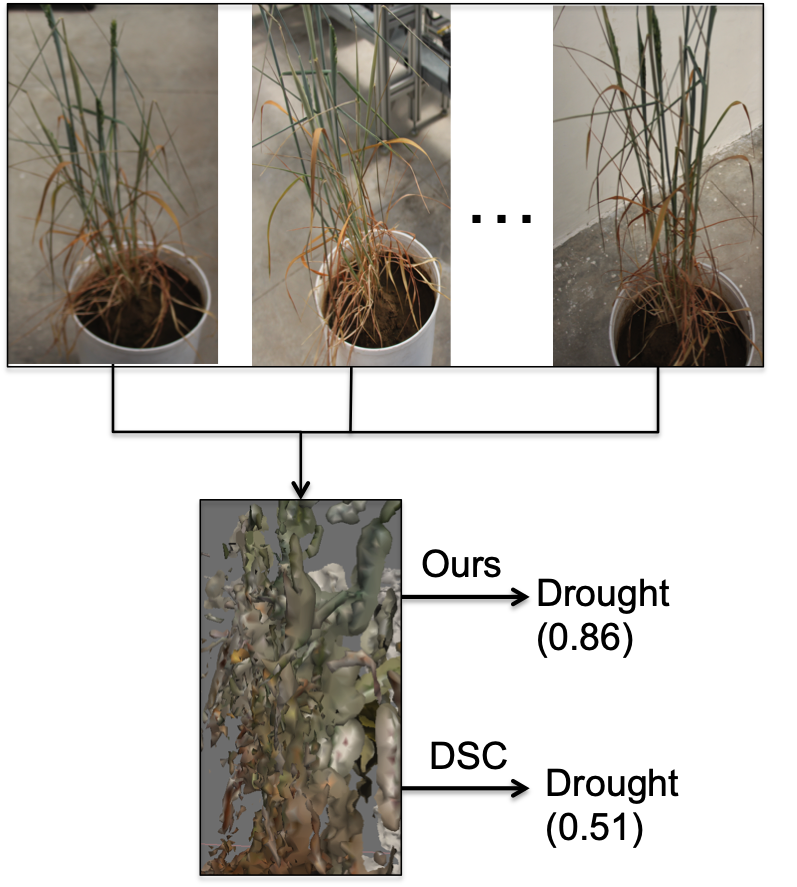} 
	\caption{
	 Qualitative results showing the classification performance on a 3D model of a plant (bottom) constructing with 2D images of a plant (top) using the proposed method (Ours) and Drought Stress Classification (DSC) \cite{srivastava2017drought}. The numbers in the bracket indicate the classification scores from the respective methods. 
	}
	\label{fig:drought}
\end{figure}

In this setting, the objective is to predict whether the input plant model is suffering drought or not. Visual identification of drought stress experienced by a plant in a non-invasive manner is an active area of research \cite{singh2016machine}. The characteristics exhibited by leaves of the plants vary significantly among plant phenotypes and genotypes which makes it difficult to develop general purpose hand-crafted descriptors for this task. To overcome this, Srivastava \etal \cite{srivastava2017drought} proposed to learn descriptors to distinguish between 3D models of drought stress experiencing plants and healthy plants. Their feature extraction pipeline involves first learning a global descriptor of a 3D plant model using PointNet, followed by aggregation of global descriptors with point features to obtain the final descriptor. However, instead of using the point features obtained from PointNet, we provide as input the features obtained from pre-trained model of 3DPatch Siamese (MMCL) which is fine-tuned on the training set of \cite{srivastava2017drought}. The results are shown in Table \ref{tab:drought}. It can be seen that the proposed architecture outperformed the best performing model of \cite{srivastava2017drought} by $\sim4$\% while having a similar computation time. 

Figure \ref{fig:drought} shows qualitative results on drought classification. The figure shows a few images of the plant from which a 3D model is constructed. The plant is in mid stages of drought as indicated by presence of both green and dry leaves. The drought stress classification technique of \cite{srivastava2017drought} which is based on optimizing global descriptors followed by local descriptors, classifies the input 3D model as suffering drought with a confidence of $51$\%. On the other hand, the proposed technique classifies the input 3D model as suffering from drought with $86$\% confidence. This shows that while both the techniques are able to classify correctly, the proposed method provides higher confidence indicating that it is able to discriminate between the variation in the local structures characterizing the drought.

\begin{table}[]
	\centering
	\caption{Classification Accuracy in 3D and Feature Computation Time}
	\footnotesize
	\label{tab:drought}
	\scalebox{0.8}{
	\begin{tabular}{|l|r|r|}
		\hline
		\multicolumn{1}{|c|}{\textbf{Descriptor}} & \multicolumn{1}{c|}{\textbf{\begin{tabular}[c]{@{}c@{}}Accuracy \\ (\%)\end{tabular}}} & \multicolumn{1}{c|}{\textbf{\begin{tabular}[c]{@{}c@{}}Computation Time (sec)\\ {[}Average Per Model{]}\end{tabular}}} \\ \hline
		SHOT (FV)                                 & 76.0                                                                                   & 5.3                                                                                                                    \\ \hline
		SHOT (BoVW)                               & 74.2                                                                                   & 6.8                                                                                                                    \\ \hline
		RoPS (FV)                                 & 77.2                                                                                   & 4.9                                                                                                                    \\ \hline
		RoPS (BoVW)                               & 75.4                                                                                   & 5.2                                                                                                                    \\ \hline
		FPFH (FV)                                 & 73.3                                                                                   & 3.9                                                                                                                    \\ \hline
		FPFH (BoVW)                               & 72.1                                                                                   & 4.3                                                                                                                    \\ \hline
		PointNet (Global)                         & 65.4                                                                                   & 2.3                                                                                                           \\ \hline
		PointNet (Aggregation)                    & 67.2                                                                                   & 2.36                                                                                                                   \\ \hline
		Fine tuned PointNet (Global)              & 76.3                                                                                   & 2.4                                                                                                                    \\ \hline
		Fine tuned PointNet (Aggregation)         & 79.1                                                                         & 2.5                                                                                                                    \\ \hline
		Fine tuned PointNet (Ours)         & \textbf{83.4}                                                                         & 2.5                                                                                                                    \\ \hline
	\end{tabular}}
\end{table}
\section{Conclusion}\label{sec:conclusion}
In this paper, we proposed a novel local 3d descriptor based on recent advances in 3d deep networks. We directly processed the input 3D patches using a deep network where the learning is made discriminative using a siamese network with a multi-margin contrastive loss. With exhaustive experiments we showed that the proposed technique outperforms competing methods. We also showed that the method is able to robustly encode the available information even when the mesh resolution is reduced, where the performance of the other methods rapidly degrades. This strengthens our hypothesis that learning directly from point cloud representations allows learning stronger and more robust descriptors. Moreover, the generalization ability and uniqueness of the proposed architecture was demonstrated with applications on classification, retrieval, binarization, object discovery and drought stress classification in plants. The experiments on classification and retrieval demonstrated that the proposed descriptor can be effectively used as a drop-in replacement for existing 3D local descriptors. On the other hand, binarization did not result in significant loss for deep learning based descriptors indicating that there is scope for learning more compact descriptors without any significant loss in robustness. Finally, we showed that the proposed descriptor can be successfully applied to real-world problem by achieving state-of-the-art results for object discovery and drought stress classification. 

\bibliographystyle{elsarticle-num}
\bibliography{IEEEfull}

\end{document}